\documentclass[letterpaper, 10 pt, conference]{ieeeconf}  

\IEEEoverridecommandlockouts                              

\usepackage{graphics} 
\usepackage{epsfig} 
\usepackage{amsmath} 
\usepackage{amssymb}  
\usepackage{float}
\usepackage{physics}
\usepackage{siunitx}
\AtBeginDocument{\RenewCommandCopy\qty\SI}
\usepackage{microtype}

\usepackage{blindtext}
\usepackage{hyperref}
\hypersetup{colorlinks, 
            linkcolor=blue,
            citecolor=blue,
            urlcolor=red}  

\usepackage{lipsum}
\usepackage{soul}
\usepackage[dvipsnames]{xcolor}

\usepackage{array}
\newcolumntype{P}[1]{>{\centering\arraybackslash}p{#1}}
\newcolumntype{L}[1]{>{\arraybackslash}p{#1}}
\usepackage{multirow}
\usepackage{subcaption}
\usepackage{dirtree}
\usepackage{balance}
\usepackage{csquotes}
\usepackage{acronym}
\usepackage{microtype}
\usepackage[style=ieee, url=false, sorting=none, doi=false, isbn=false, eprint=false, maxnames=1, minnames=1, dashed=false]{biblatex}
\usepackage{siunitx}
\usepackage{booktabs}
\usepackage{tabularx}
\usepackage{acronym}           
\usepackage{bm}
\usepackage{acronym}
\newacro{FEM}{finite element method}

\title{\LARGE \bf
When Obstacles Bend: Modeling Vegetation Deformation in the context of Field Robotics}

\author{Zaar Khizar$^{1}$, Tom Montagnon$^{2}$, Roland Lenain$^{2}$, Romuald Aufrère$^{3}$ and Johann Laconte$^{2}$
\thanks{$^{1}$ LIMOS, Université Clermont Auvergne,  Clermont Auvergne INP, CNRS,  F-63000 Clermont-Ferrand, France}%
\thanks{$^{2}$ Université Clermont Auvergne, INRAE, UR TSCF, 63000, Clermont-Ferrand, France }%
\thanks{$^{3}$ Institut Pascal, Université Clermont Auvergne, Clermont Auvergne INP, CNRS,  F-63000 Clermont-Ferrand, France}%
}

\begin{document}
\hyphenpenalty=10000
\exhyphenpenalty=10000

\maketitle

\begin{abstract}\label{sec:abstract}
%
Autonomous robots operating in natural environments must often interact with vegetation rather than simply avoid it.
In this context, traversability is typically defined from the robot’s perspective, by measuring how a specific platform responds when moving through the environment.
While practical, this viewpoint entangles the assessment of the environment with the robot’s own dynamics, making the resulting characterization difficult to transfer across different platforms.
More importantly, it does not directly reflect the properties of the vegetation itself, which are the true source of interaction and potential damage in applications such as agriculture and environmental monitoring.
To address this limitation, we propose to characterize vegetation through its intrinsic mechanical properties, independently of any specific robot.
By combining deformation measurements with contact force data, we estimate the underlying mechanical parameters and reconstruct the vegetation’s response to interaction.
This enables vegetation-aware navigation based on intrinsic environmental properties rather than platform-dependent metrics.
\end{abstract}

\section{INTRODUCTION}\label{sec:introduction}
\begin{figure}[t]
    \centering
    \includegraphics[width=\linewidth]{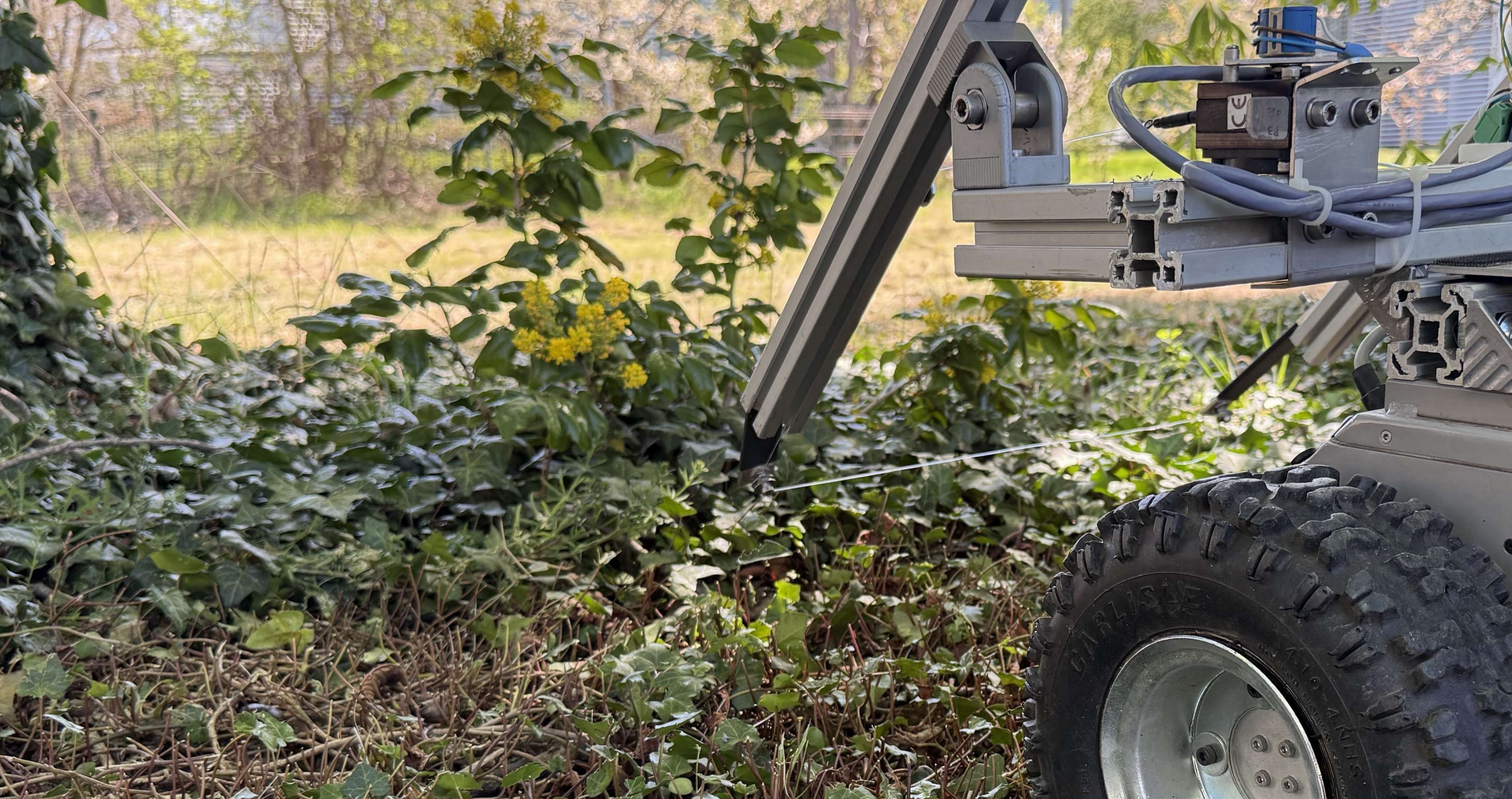}
    \caption{A mobile robot facing a patch of mixed vegetation, deciding whether to push through it or go around. From appearance alone, the ivy, brambles, and woody stems are difficult to distinguish, yet they differ in how much they resist being pushed through.}
    \label{fig:cover_photo}
    \vspace{-2.2em}
\end{figure}

\enlargethispage{\baselineskip}
Autonomous robots increasingly operate in vegetative environments where vegetation is not simply an obstacle to avoid but a deformable medium through which motion and interaction is often necessary.
Modeling the environment before acting in it is a foundational principle in robotics: planning and control need a known state of the world, not one reacted to as encountered.
For vegetation, however, the property that matters most is mechanical rather than purely geometric: it occupies the same region of space as a rigid obstacle would, but whether a robot can and should push through it, deflect it, or must circumvent it depends on the plant's mechanical behavior, not merely its visible size or shape, as shown in \autoref{fig:cover_photo}.
This distinction matters beyond robotics: a harvesting robot that cannot tell a stiff stalk from a compliant one, for example a strawberry runner from its fruit-bearing stem, risks damaging the crop.
Conventional representations capture where matter is, not how it behaves mechanically.
Lacking a way to measure this directly, recent methods estimate traversal difficulty from how a specific robot fared, not from the vegetation's own properties.
Recent surveys \cite{borges2022survey, beycimen2023comprehensive} group these vision-, geometry-, and hybrid-based methods together: each estimates a cost tied to the robot that produced it, not to the terrain itself, so locomotion-based estimates differ across platforms traversing identical terrain \cite{eder2023traversability}.
This is a structural limitation rather than one of insufficient data: the terrain itself is never described independently from the robot that traversed it, so the model must be re-estimated whenever the robot, sensor mounting, or control policy changes.

An alternative is to describe the vegetation as a mechanical structure with properties that exist independently of any robot.
This is a different kind of problem, because a plant's behavior is not adequately described by a single value: it is a continuous structure whose resistance to bending changes along its length, and observing how it deforms under contact constrains this behavior only partially.
Recovering a physically grounded description therefore requires an underlying mechanical model relating what is observed to what the plant's structure actually is, together with a sensing approach capable of capturing that interaction.

\enlargethispage{\baselineskip}
This paper addresses that gap by grounding vegetation mechanics in rod theory \cite{goriely2006mechanics, guillon2012new}, a single foundation that captures two distinct modes of a plant's motion: smooth bending along its length, and near-rigid rotation about a compliant base.
Combined with force and shape measurements, this foundation recovers each directly from the interaction itself, independently of the robot or sensor used to collect it: a distributed stiffness profile for slower, detailed tasks such as crop-lodging assessment and forestry understory characterization, and a single lumped stiffness for real-time traversability decisions.
The main contributions of this paper are: 
(i) A unified rod-theoretic framework for modeling vegetation stiffness;
(ii)  a state estimation pipeline recovering the underlying mechanical parameters of the vegetation; and
(iii) an experimental evaluation of the two proposed mechanical models, highlighting their respective strengths and limitations for robotic applications.

\section{Related Work}\label{sec:lit_review}

The dominant line of work on robot interaction with vegetation treats it as a traversability problem: environments are represented with standard terrain models, annotated with a scalar cost derived from geometry \cite{papadakis2013survey} or learned from robot performance signals such as tracking error \cite{gasparino2024wayfaster} or vibration \cite{sathyamoorthy2022terrapn}.
However, these methods characterize how a specific robot fares against vegetation and not what the vegetation itself mechanically is; our concern is not with defining a traversability cost, but with recovering a mechanical description of the plant that a cost, or any other downstream task, could later be built on.

A smaller body of work models obstacle deformation directly, instead of inferring a cost from it: \textcite{frank2014learning} do so with a finite element formulation, computing an interaction cost for planning and showing that the underlying material parameters can themselves be deduced from force and visual observations.
\textcite{arriola2020modeling} note that while \ac{FEM} is physically faithful, it requires offline meshing and controlled identification experiments, and has been validated mainly on simple, roughly isotropic materials such as foam and cloth, not the branching, tapered, anisotropic geometry of a vegetation stem, a trade-off that motivates the lower-dimensional continuum model, tailored to slender, tapering structures, used here.

Representing a stem as a Kirchhoff rod, a one-dimensional elastic continuum with inextensibility and cross-sectional planarity constraints, is established practice in both robotics \cite{rucker2011statics} and plant biomechanics \cite{goriely2006mechanics}, justifying its use here over a heavier representation such as a full \ac{FEM} mesh. \textcite{deng2024gazebo} apply this same rod formalism to plant-robot contact, but calibrate its stiffness offline against reference video for use inside a simulator, so their result is a simulated rendering rather than a physically validated property of a real stem.
What remains open is how to identify such a rod's parameters from a robot's own sensors during ordinary interaction, as opposed to from simulation calibration or a separate laboratory test.

The natural candidate for that identification is direct contact sensing, which recovers mechanical information geometry or appearance cannot, though each method below leaves at least one limitation open.
\textcite{ordonez2018modeling, ordonez2020characterization} fit each stem's rotational stiffness and damping from manually displaced stems in the lab and confirm in the field, via manipulator probing, that the same parameters predict traversal energy. \textcite{haddeler2022traversability} similarly mount a probing arm on a quadruped for a collapsibility metric; all three require a dedicated probing action outside ordinary driving.
\textcite{ehrlich2026tactile} instead convert ordinary contact into a discrete, timestamped force proxy rather than a continuous signal, and \textcite{xuan2024environmental} estimate a single scalar beam stiffness during ordinary traversal, collapsing the response to one number in place of a distributed profile.
The wire-based sensor of \textcite{khizar2025feeling} removes the probing requirement and reports a force signal, but without relating it to the plant's own structure, leaving the distributed-profile limitation open, which the present paper closes.

The two lines of work closest to the present paper are \textcite{khizar2025feeling}, who report a continuous force signal from ordinary contact but no mechanical model, and \textcite{deng2024gazebo}, introduced above.
This paper combines the sensing of the former with the rod-theoretic representation of the latter, posing the stem's distributed flexural rigidity as an inverse problem solved directly from force and shape measured on a real robot during ordinary traversal.
The result is a mechanical descriptor characterized independently of the robot, sensor, or control policy used to collect it.
\begin{figure*}[t!]
    \centering
    \vspace{-4em}
    \includegraphics[width=\linewidth]{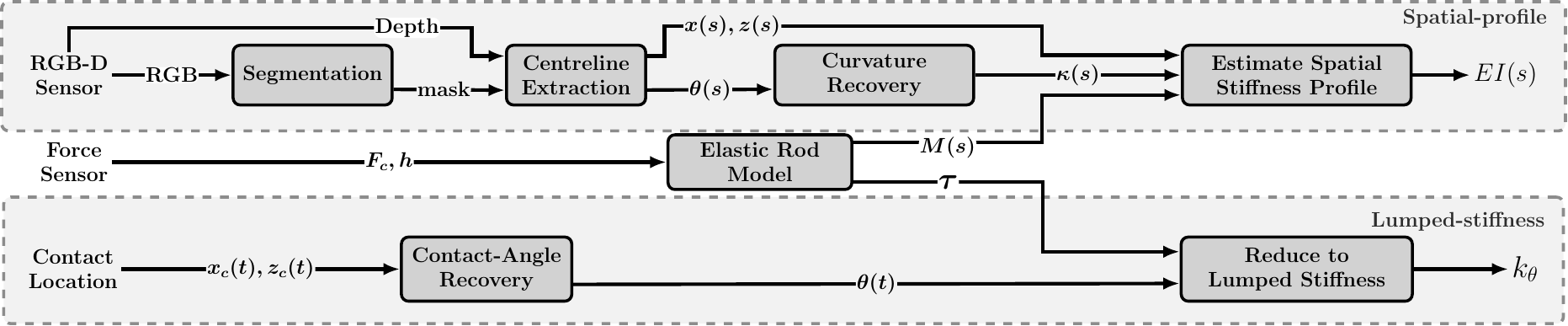}
    \caption{Overview of the rod-theoretic modeling pipeline.
    The RGB-D stream is segmented and reduced to a centerline and curvature $\kappa(s)$, while the contact loading $F_c$ and initial contact height $h$ are supplied to the Elastic Rod Model, which returns the internal moment $M(s)$, and the base torque $\tau$.
    The spatial-profile branch (top) combines $\kappa(s)$ and $M(s)$ into the flexural-rigidity profile $EI(s)$; the lumped-stiffness branch (bottom) recovers the deflection angle $\theta(t)$ from the contact location $(x_c(t), z_c(t))$ and combines it with $\tau$ to give the rotational stiffness $k_\theta$.
    $EI(s)$ and $k_\theta$ are two independent outputs of the same underlying model, and the choice between them depends on the intended downstream application.}
    \label{fig:global_pipeline}
    \vspace{-1em}
\end{figure*}
\section{Rod-Theoretic Modeling of Vegetation}\label{sec:theory}
\begin{figure}[h!]
        \centering
        \begin{subfigure}{0.49\linewidth}
            \includegraphics[width=\linewidth]{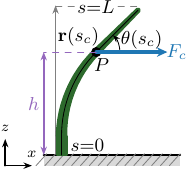}
            \caption{}
            \label{fig:kirchhoff_rod}
        \end{subfigure}
        \begin{subfigure}{0.49\linewidth}
            \includegraphics[width=\linewidth]{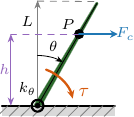}
            \caption{}
            \label{fig:spring_rod}
        \end{subfigure}
        \vspace{-2em}
        \caption{Both panels show a tapered stem of length $L$, clamped at its base, under a horizontal contact force $F_c$ at height $h$, point $P$.
        (a) The Kirchhoff rod bends continuously, with contact point $P=\mathbf{r}(s_c)$ and tangent angle $\theta(s)$ varying along the rod.
        (b) The spring rod reduces the stem to a rigid link on an equivalent spring $k_\theta$, representing its distributed compliance below $P$: the same force gives one deflection angle $\theta$ and base torque $\tau$.}
        \label{fig:rods}
        \vspace{-1.25em}
    \end{figure}

When a robot advances through vegetation, it pushes against individual stems, and the resulting interaction reveals how stiff each stem is and how that stiffness is distributed along its length, information beyond what image or geometry alone can provide.
This stiffness is the stem's flexural rigidity, $EI(s)$, the product of the material's Young's modulus $E$ and the cross-section's second moment of area $I(s)$ at arc length $s$; it governs the stem's entire mechanical response and, unlike visible geometry, cannot be read off an image or assumed constant across a species.
Recovering it, rather than assuming it or measuring it destructively, is the central problem this section addresses, using a planar Kirchhoff rod paired with a robot-mounted force sensor that measures the push as the robot advances.
How much of the flexural rigidity $EI(s)$ a task needs varies: crop-lodging assessment and forestry characterization need its full spatial detail and can justify a camera to get it, while real-time traversability decisions need only the stem's aggregate resistance from the force sensor alone.
This section therefore derives two outputs from the same rod: the full spatial profile $EI(s)$, recovered from the stem's observed shape and contact force, and a lumped rotational stiffness $k_\theta$, a torque-based reduction of $EI(s)$ recovered from the contact force alone.
\autoref{fig:global_pipeline} summarizes the resulting pipeline, and \autoref{fig:rods} shows the two rod idealizations behind each output: a distributed rod clamped at the base (\autoref{fig:kirchhoff_rod}) for the spatial profile, following the Kirchhoff-rod treatment of plant stems \cite{goriely2006mechanics}, and a rigid link on an equivalent rotational spring (\autoref{fig:spring_rod}) for the lumped stiffness, following the base rotational-stiffness idealization used to characterize individual stems \cite{ordonez2018modeling}.
\subsection{Problem Setup and Assumptions}\label{sec:assumptions}

A vegetation stem, rooted in the ground, has arc length $L$ measured from base to tip.
The robot carries a force sensor at fixed height $h$ that contacts and bends the stem as it advances, outputting a force $F_c$; the model requires only this force, so any force or torque sensor with this output can be used.
This paper uses the wire-based sensor of \cite{khizar2025feeling} rather than a commercial force or torque sensor, since it is field-deployable, inexpensive, compliant, and low-noise at stem-contact magnitudes. As shown in \autoref{sec:lumped_reduction}, its elongation also yields the deflection angle for the lumped-stiffness branch without a camera.
Where the vision pipeline of \autoref{sec:shape_extraction} is used instead, a stationary RGB-D camera observes the stem side-on, with the bending plane parallel to the image plane.

Since a robot can approach and push a stem from any direction, the loading has no single preferred azimuth around the stem, motivating two simplifications used throughout this section: reducing the three-dimensional bending problem to a single plane containing the load, and treating the stem's cross-section as circular.

The stem is modeled as a planar Kirchhoff rod, a one-dimensional elastic continuum whose configuration is fully described by its centerline $\mathbf{r}(s) = (x(s), z(s))$ and tangent angle $\theta(s)$, $s \in [0, L]$, resting on six assumptions referenced by label throughout the section:

\begin{enumerate}
    \item[A1] \textit{Planarity and static camera.} Bending occurs in a single vertical plane, coincident with a stationary image plane when a camera is used.
    \item[A2] \textit{Quasi-static loading.} Inertial and damping terms are neglected; the rod is in static equilibrium at every instant, since the robot advances slowly relative to the stem's natural response time.
    \item[A3] \textit{Inextensibility and unshearability.} The rod neither stretches nor shears, only bends.
    \item[A4] \textit{Linear elastic constitutive law.} The bending moment is linear in the curvature deviation from a rest configuration, with proportionality constant $EI(s)$.
    \item[A5] \textit{Single point contact.} Contact with a real plant is approximated as a single concentrated transverse force at arc length $s_c(t)$, using an equivalent single rod, since the sensor's contact width is small relative to the stem length it engages.
    \item[A6] \textit{Boundary conditions.} The base is clamped, and vertical, so $s=0$ is fixed with unstressed tangent vertical; the tip at $s=L$ is free of external force or moment.
\end{enumerate}

\subsection{Rod Kinematics, Equilibrium, and Flexural-Rigidity Estimation}\label{sec:rod_model}
    \subsubsection{Kinematics, Equilibrium, and Contact Moment}

        The Kirchhoff rod used here is the planar (A1), unshearable (A3), no-twist specialization of the general Cosserat rod \cite{antman2005nonlinear}.
        Planarity and unshearability reduce the rod's configuration to a single tangent angle $\theta(s)$, measured from the horizontal as in \autoref{fig:kirchhoff_rod}, and inextensibility (A3) fixes this tangent to unit norm under the resulting arc-length parametrization:
        \begin{equation}\label{eq:tangent}
            \begin{split}
            \mathbf{r}'(s) &= \big(\cos\theta(s),\, \sin\theta(s)\big), \\
            x'(s)&=\cos\theta(s),\ z'(s)=\sin\theta(s).
            \end{split}
        \end{equation}
        Here and throughout this section, the $(\cdot)^{\prime}$ denotes differentiation with respect to arc length $s$.
        The curvature is the rotation rate of this angle, $\kappa(s)=\theta'(s)$, from \eqref{eq:tangent}.

        Let $\mathbf{n}(s)$ and $M(s)$ denote the internal force and moment exerted by the tip-side of the rod on the base-side at arc length $s$, and $\mathbf{f}(s)$ the external force per unit arc length.
        Newton's third law and a moment balance about $\mathbf{r}(s)$ give the standard planar Kirchhoff equilibrium equations,
        \begin{align}
        \mathbf{n}'(s) + \mathbf{f}(s) &= \mathbf{0}, \label{eq:force_balance} \\
        M'(s) &= n_z(s)\cos\theta(s) - n_x(s)\sin\theta(s), \label{eq:moment_balance_scalar}
        \end{align}
        with $\mathbf{n}=(n_x,0,n_z)$ and no inertial term, by the quasi-static assumption (A2).
        For a linearly elastic rod (A4), the bending strain energy per unit length is quadratic in curvature,
        \begin{equation}\label{eq:energy_density}
            \begin{split}
                u(s) &= \tfrac{1}{2}EI(s)\,\kappa(s)^2, \\
                U &= \int_0^L u(s)\,ds,
            \end{split}
        \end{equation}
        and its conjugate moment, $M(s) := \partial u/\partial\kappa$, gives the constitutive law
        \begin{equation}\label{eq:constitutive}
            M(s) = EI(s)\,\kappa(s).
        \end{equation}
        \eqref{eq:force_balance}, \eqref{eq:moment_balance_scalar}, and \eqref{eq:constitutive} constitute the Elastic Rod Model block of \autoref{fig:global_pipeline}.
        Together with $\kappa(s)=\theta'(s)$, these form a boundary value problem in $\theta$: A6 fixes $\theta$ at the base and $M$ at the tip.
        What remains is an explicit expression for $M(s)$ in terms of the measured contact force $F_c$, supplied next by the point-contact geometry.
        The sensor applies a single horizontal point force of magnitude $F_c$ and direction $\hat{s}\in\{-1,+1\}$ at arc length $s_c$ (A5), as illustrated in \autoref{fig:kirchhoff_rod}.
        Because the tip is free, $\mathbf{n}(L)=\mathbf{0}$, integrating \eqref{eq:force_balance} across $s_c$ shows that the internal force is constant on each side of the contact point: $(n_x,n_z)=(\hat{s}F_c,0)$ below it, and $\mathbf{n}=\mathbf{0}$ above it, so $M(s)=0$ for $s\in(s_c,L]$ as well.
        Substituting this into \eqref{eq:moment_balance_scalar} and integrating gives the moment in closed form,
        \begin{equation}\label{eq:M_determinate}
            M(s) =
            \begin{cases}
                \hat{s}F_c\big(h-z(s)\big), & 0\le s\le s_c,\\
                0, & s_c<s\le L,
            \end{cases}
        \end{equation}
        using $z(s_c){=}h$, the contact height.
        $M(s)$ is therefore fixed by the measured force $F_c$ and the observed shape $z(s)$ alone, independent of the flexural rigidity $EI(s)$. Recovering $EI(s)$ from \eqref{eq:constitutive} then requires only the curvature $\kappa(s)$.
        \autoref{sec:shape_extraction} obtains $\kappa(s)$ from RGB-D observations, which is then combined with \eqref{eq:M_determinate} to estimate $EI(s)$, corresponding to the Estimate Spatial Stiffness Profile block of \autoref{fig:global_pipeline}.

    \subsubsection{Direct Estimation of \texorpdfstring{$EI(s)$}{EI(s)}}

        The static relations of \eqref{eq:M_determinate} and \eqref{eq:constitutive} hold at every instant, so this section evaluates them frame by frame: the observed shape $\widehat{z}(t,s)$, moment $\widehat{M}(t,s)$, and curvature $\widehat{\kappa}(t,s)$ each carry a frame index $t$ from the shape sequence of \autoref{sec:shape_extraction}, in addition to arc length $s$.
        Combined with the constitutive law of \eqref{eq:constitutive}, $EI(s)$ can be recovered directly at every observed arc-length bin and frame,
        \begin{equation}\label{eq:direct_estimator}
        \widehat{EI}_{\mathrm{sample}}(t,s) = \frac{\widehat{M}(t,s)}{\widehat{\kappa}(t,s)},
        \end{equation}
        This estimator is model-free in the sense that it makes no assumption about how $EI(s)$ varies along the stem, but is ill-conditioned near zero curvature, close to the base or early in a push, since a fixed measurement noise floor in the denominator produces unbounded relative error as $\widehat{\kappa}\to0$.
        Averaging out this noise, without trusting any single noisy bin, requires fitting a low-dimensional model of how $EI$ varies along the stem, developed next.

    \subsubsection{Flexural-Rigidity Parametrization and Fitting}

        Under the circular cross-section adopted in \autoref{sec:assumptions}, the second moment of area is $I(s) = \pi D(s)^4/64$, independent of loading direction \cite{banyai2025biomechanics}.
        The diameter $D(s)$ tapers approximately as a power law along the stem \cite{niklas1992plant}, as given in \autoref{eq:diameter_taper}.
        \begin{equation}\label{eq:diameter_taper}
            D(s) = d_0\Big(1-\frac{s}{L}\Big)^{p}.
        \end{equation}
        Here $d_0$ is the base diameter and $p$ is the taper exponent.
        Substituting into $I(s)$ and multiplying by the Young's modulus $E$ gives
        \begin{equation}\label{eq:EI_taper}
            EI(s) = EI_0\Big(1-\frac{s}{L}\Big)^{4p}, \quad EI_0 := \frac{\pi E d_0^4}{64},
        \end{equation}
        where $EI_0$ is the base flexural rigidity, $EI(0)$.
        $EI(s)$ is therefore fully characterized by two parameters, the base rigidity $EI_0$ and the taper exponent $p$: estimating the profile reduces to estimating these two quantities.

        A single push identifies $p$ poorly: it engages only a short span of the stem, starting near arc length $s\approx h$, where contact begins while deflection is still small, and extending to $s_c$, which grows only modestly beyond that as the push proceeds.
        Over this short span, a stiffer base with a steeper taper can approximately reproduce the same moment-curvature relation as a softer base with a gentler taper, so the two parameters of \eqref{eq:EI_taper} are only weakly distinguishable from one trial.
        This ambiguity is resolved by pooling the direct estimate of \eqref{eq:direct_estimator} across contacts made at several heights $h$, since $s_c$ then spans a much larger portion of $[0,L]$ than any single push can reach.
        Substituting \eqref{eq:tangent}, \eqref{eq:constitutive}, and \eqref{eq:EI_taper} into the equilibrium equations, \eqref{eq:force_balance} and \eqref{eq:moment_balance_scalar}, gives the rod state $\bm{\xi}(s)=[\,x(s),\ z(s),\ \theta(s)\,]^{T}$ as the initial value problem
        \begin{equation}\label{eq:ivp}
            \frac{d}{ds}\bm{\xi}(s)
            =
            \begin{bmatrix} \cos\theta \\ \sin\theta \\ \hat{s}F_c(h-z)/EI(s) \end{bmatrix},
            \qquad z(0)=0,
        \end{equation}
        with $\theta'=0$ once $z\ge h$, per \eqref{eq:M_determinate}.

        The taper exponent $p$ is fit first, as $p=b/4$ where $a$ and $b$ are the intercept, $\ln{EI_0}$, and slope of a weighted log-linear fit to \eqref{eq:direct_estimator}, pooled across contact heights $h$ and binned by arc length $s_i$,
        \begin{equation}\label{eq:p_fit}
            (a,\,b) = \operatorname*{argmin}_{a,\,b}\ \sum_i w_i\Big(\ln\widehat{EI}(s_i) - a - b\ln(1-s_i/L)\Big)^2,
        \end{equation}
        with $w_i$ the inverse square of bin $i$'s relative standard error, so noisy bins near $\widehat{\kappa}\to0$ contribute less; the intercept $a$ is not used further, since it inherits the same near-base sensitivity as \eqref{eq:direct_estimator}.
        With $p$ fixed at this value, $EI_0$ is instead recovered from the shape \eqref{eq:ivp} predicts: with $\mathbf{r}(s_c;EI_0)$ the position given by integrating \eqref{eq:ivp} to $s_c$, and $\mathbf{r}_{\mathrm{obs}}(t,s_c)$ the observed centerline position at frame $t$'s contact point,
        \begin{equation}\label{eq:EI0_fit}
            EI_0 = \operatorname*{argmin}_{EI_0}\ \sum_t \big\|\mathbf{r}(s_c;EI_0) - \mathbf{r}_{\mathrm{obs}}(t,s_c)\big\|^2,
        \end{equation}
        the resulting position residual, summed over all engaged frames $t$, avoiding the curvature-based ratio of \eqref{eq:direct_estimator}, which is least reliable exactly at $s=0$ where $EI_0$ is defined.

        As such, estimating the profile's intrinsic parameters, $EI_0$ and $p$, requires both \eqref{eq:p_fit}, which rests on the direct estimate of \eqref{eq:direct_estimator}, and \eqref{eq:EI0_fit}, which rests on \eqref{eq:ivp}; both in turn depend on \eqref{eq:M_determinate} for the moment and \eqref{eq:EI_taper} for the stiffness model.
        This is in contrast to \eqref{eq:direct_estimator}, which recovers $EI(s)$ pointwise without assuming a parametric form, relying only on \eqref{eq:M_determinate}.

    \subsection{Shape Extraction from RGB-D Observations}\label{sec:shape_extraction}

        The estimators of \autoref{sec:rod_model} each require a different observed quantity at every instant: the direct estimator needs the curvature $\kappa(s)$, per \eqref{eq:direct_estimator}, and the model-based estimator needs the full position centerline, per \eqref{eq:ivp}.
        Both are obtained from the same vision pipeline.

        Each frame is segmented (Segmentation block, \autoref{fig:global_pipeline}) and the stem centerline is extracted as a path between its base and tip (Centerline Extraction block, \autoref{fig:global_pipeline}), then fit with a smoothing spline resampled at equal arc-length increments, using the depth channel to convert pixel coordinates into a metric arc length.
        This spline gives the position centerline $(x(s),z(s))$ directly.
        Differentiating raw, per-frame noise degrades sharply with sampling density; the tangent angle field $\theta(s)$ is therefore further smoothed by Gaussian process regression, with a covariance chosen so that sample paths remain differentiable enough for the curvature $\kappa(s)=\theta'(s)$ of \eqref{eq:tangent} to exist and stay finite, without over-smoothing the underlying shape (Curvature Recovery block, \autoref{fig:global_pipeline}).
        The resulting centerline and curvature, together with the measured contact force $F_c$, are passed to the $EI(s)$ estimators, the direct estimator of \eqref{eq:direct_estimator} and the model-based estimator of \eqref{eq:ivp}, as shown in \autoref{fig:global_pipeline}.

    \subsection{Lumped-Parameter Reduction}\label{sec:lumped_reduction}

        While \autoref{sec:rod_model} recovers the flexural-rigidity profile $EI(s)$, many downstream tasks do not need this much detail: field navigation typically needs only the peak force a stem will exert and the angle at which it yields \cite{ordonez2020characterization}.
        This is the lumped branch of \autoref{fig:global_pipeline}, and it is what makes the model usable in real time without the vision pipeline of \autoref{sec:shape_extraction}.
        This subsection reduces the rod to a single scalar stiffness.
        It relates the contact force $F_c$ to a deflection, either a rotation $\theta$ or a horizontal displacement $\Delta x$, at the contact point.
        \autoref{fig:spring_rod} shows this small-deflection reduction.
        The deflection it requires, $\theta$ or $\Delta x$, is then recovered from the contact location alone, without a camera.

        \subsubsection{Small-Deflection Reduction}

            The spring rod of \autoref{fig:spring_rod} reduces the whole stem to a rigid link on a single rotational spring.
            This spring represents the stem's own distributed compliance below the contact point.
            It stores energy $U = \tfrac12 k_\theta \theta^2$ for a rotation $\theta$, with $k_\theta$ its rotational stiffness.
            Differentiating gives its restoring torque, $\partial U/\partial\theta = k_\theta\theta$, the lumped analogue of the constitutive law \eqref{eq:constitutive}.
            Quasi-static equilibrium (A2) balances this restoring torque against the applied torque $\tau$, from the contact force $F_c$ at height $h$,
            \begin{equation}\label{eq:torque_balance}
                \tau = F_c\,h = k_\theta\,\theta.
            \end{equation}

            What remains is an explicit expression for $k_\theta$ itself, in terms of the stem's own flexural rigidity $EI(s)$.
            In the small-deflection limit, arc length is approximated by horizontal position, $s\approx x$, so the contact moment of \eqref{eq:M_determinate} varies linearly over the loaded span, producing a curvature that varies with $s$.
            Introducing an auxiliary moment $M_a$ at $s=h$ and substituting $M(s)=F_c(h-s)+M_a$ into the bending energy of \eqref{eq:energy_density} via the constitutive law \eqref{eq:constitutive} gives the complementary energy
            \begin{equation}\label{eq:Ustar}
                U^*(F_c,M_a) = \int_0^h \frac{\big(F_c(h-s)+M_a\big)^2}{2EI(s)}\,ds,
            \end{equation}
            Differentiating this with respect to $M_a$ and $F_c$, at $M_a=0$, gives the rotational and translational stiffnesses at once,
            \begin{align}
                k_\theta(h) &= \left[\frac1h\int_0^h \frac{h-s}{EI(s)}\,ds\right]^{-1}, \label{eq:kdist}\\
                k_x(h) &= \left[\int_0^h \frac{(h-s)^2}{EI(s)}\,ds\right]^{-1}, \label{eq:kx}
            \end{align}
            with $k_x$ the translational stiffness relating the contact force $F_c$ to a horizontal displacement $\Delta x$ at height $h$.
            For a constant $EI$, these reduce to $k_\theta=2EI/h$ and $k_x=3EI/h^3$, the standard cantilever rotation and tip-deflection stiffnesses under an end point load \cite{GereJamesM2013Mom}.
            The rotational stiffness $k_\theta$ enters the torque balance of \eqref{eq:torque_balance}, while the translational stiffness $k_x$ enters the independent, parallel force balance
            \begin{equation}\label{eq:kx_balance}
                F_c = k_x(h)\,\Delta x(h).
            \end{equation}
            Which to use depends on whether the downstream task needs a restoring torque and rotation or a restoring force and displacement.
            Both stiffnesses are otherwise small-deflection results; where larger rotations make this linearization inaccurate and the stem shape is observed, the equivalent secant stiffness $F_ch/\theta(h)$ can instead be read from the nonlinear elastica of \eqref{eq:ivp}.

        \subsubsection{Camera-Free Angle Recovery from the Wire Sensor}

            The wire sensor \cite{khizar2025feeling}, chosen for the reasons noted in \autoref{sec:assumptions}, consists of a single cable anchored at two fixed points on the robot, spanning its front at height $h$ with unstretched length $L_w$, and outputs only the resulting elongation of this span.

            As the robot advances and a stem pushes against the wire, the contact point is displaced laterally by $\Delta x$, stretching each half of the wire from $L_w/2$ to $\sqrt{(L_w/2)^2+\Delta x^2}$; the resulting elongation $\Delta\ell(t)$ is what the sensor measures directly.
            Specialized to contact at the wire's midpoint, inverting this elongation model gives the lateral displacement directly from the measured elongation,
            \begin{equation}\label{eq:wire_half_ext}
            \Delta x = \tfrac12\sqrt{(\Delta\ell+L_w)^2-L_w^2},
            \end{equation}
            which is exactly the $\Delta x$ needed in \eqref{eq:kx_balance}.

            Robot pose and $\Delta x$ together fix the contact location $(x_c(t), z_c(t))$ in the base-fixed frame, with $x_c$ the horizontal distance advanced since first contact and $z_c$ its height above the base.
            Together with the wire's fixed mounting height $h$, geometry then fixes $\theta$ (\autoref{fig:spring_rod}) through
            \begin{equation}\label{eq:angle_geom}
            x_c\cos\theta - h\sin\theta = z_c,
            \end{equation}
            a single trigonometric equation in $\theta$, with closed-form solution
            \begin{equation}\label{eq:angle_solution}
            \theta = 2\arctan\!\left(\frac{\sqrt{h^2+x_c^2-z_c^2}-h}{x_c+z_c}\right).
            \end{equation}
            With $\theta$ and $\Delta x$ recovered from the wire elongation $\Delta\ell(t)$ and the robot's own pose, and $\tau=F_ch$ from the measured force and known height, the lumped stiffnesses of \eqref{eq:kx_balance} and \eqref{eq:torque_balance} are fully specified without the vision pipeline of \autoref{sec:shape_extraction}.

\medskip
Both models of \autoref{fig:global_pipeline}, and both rod idealizations of \autoref{fig:rods}, ultimately estimate a material or geometric property of the stem itself, either the distributed $EI(s)$ or the aggregate $k_\theta$, neither a property of the sensor or robot that produced the measurement.
The flexural-rigidity $EI(s)$ additionally does not depend on the contact height it was measured at; the rotational stiffness $k_\theta$ does, and must be refit at each new height, as \autoref{sec:discussion} shows.

\section{Experiments and Results}\label{sec:results}

    This section evaluates the rod-theoretic framework of \autoref{sec:theory} against physical push-through data collected on two specimens.

\subsection{Experimental Setup and Vision Pipeline Implementation}\label{sec:experimental_setup}
\begin{figure}[h!]
    \centering
    \vspace{-1em}
        \begin{subfigure}{0.49\linewidth}
            \includegraphics[width=\linewidth]{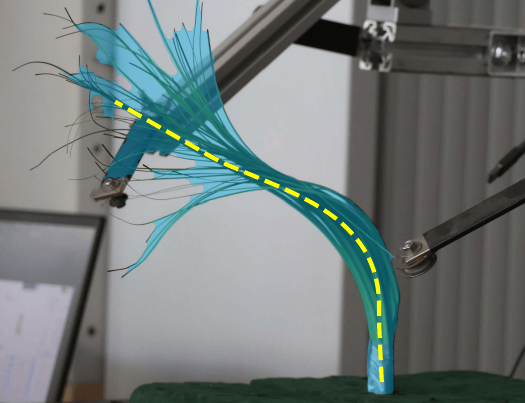}
            \caption{}
            \label{fig:setup_grass}
        \end{subfigure}
        \hfill
        \begin{subfigure}{0.49\linewidth}
            \includegraphics[width=0.81\linewidth]{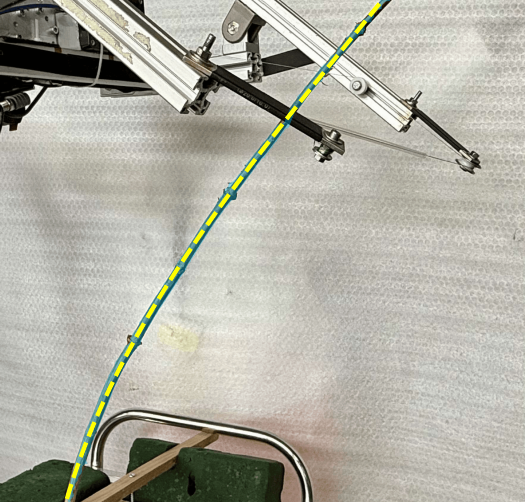}
            \caption{}
            \label{fig:setup_twig}
        \end{subfigure}
    \caption{Experimental specimens pushed by the wire sensor of \cite{khizar2025feeling}, with the segmentation mask (blue) and extracted centerline (dashed yellow) overlaid. (a) Artificial grass bush. (b) Woody twig collected from the wild.}
    \label{fig:setup}
    \vspace{-1em}
\end{figure}

This subsection describes the two vegetation specimens tested, the setup used to gather data and the implementation of the vision pipeline part of \autoref{fig:global_pipeline}.
Two specimens are used: an artificial grass bush, mimicking the geometry of a real grass tuft, and a woody twig.
A living, rooted grass plant was not used because repeated manipulation permanently changes its mechanical properties.
The artificial bush maintains consistent mechanical properties across repeated interactions, allowing statistical analysis as presented below.
However, note that the method presented in this paper applies to any deformable entity.
Results are reported for these two specimens, shown in \autoref{fig:setup}: the grass bush has total length $L=\SI{0.47}{m}$, and the twig $L=\SI{0.80}{m}$.

Both specimens were pushed from first contact to full traversal by a wire force sensor \cite{khizar2025feeling} mounted on a UR10 arm, held at a constant height throughout each push.
The vegetation was probed at different contact heights to collect data across its length.
The usable contact heights are bounded at both ends by the sensor's force range: near the base, the short lever arm combined with stiff material generates forces that are too large, while near the tip, the long lever arm combined with compliant material generates forces that are too small.
The initial contact height for the grass and twig has been respectively set to $h=\{9,11,14\}\si{\cm}$ and $h=\{45,55,65\}\si{\cm}$.

A stationary Zed2i camera recorded RGB-D video for shape extraction.
Each RGB frame is segmented with SAM2~\cite{ravi2025sam} and the mask is skeletonized by morphological thinning into an 8-connected pixel graph.
The base and tip nodes are joined by Dijkstra's shortest path to give the stem's centerline.
It is converted to metric arc length via the depth channel, fit with a cubic B-spline, and resampled at equal arc-length increments to obtain the centerline $(x(s),z(s))$ and tangent angle $\theta(s)$.
The tangent field $\theta(s)$ is smoothed by Gaussian process regression with a Mat\'ern-$3/2$ covariance, providing the differentiability required to compute the curvature $\kappa(s)$.
\subsection{Flexural-Rigidity Profile \texorpdfstring{$EI(s)$}{EI(s)}}\label{sec:ei_results}
In this section, we investigate the applicability of the Kirchhoff-rod model, as well as the added value of the taper-law parametrization over the direct measurement.
The direct estimator of \eqref{eq:direct_estimator} recovers $EI(s)$ from moment $M(s)$ and curvature $\kappa(s)$ alone, with no assumption about the functional form of $EI(s)$.
However, near the base, curvature $\kappa$ is small; as $\kappa\to0$, \eqref{eq:direct_estimator} shows the signal-to-noise ratio worsening, and the data become unusable in that range.
As such, the taper law of \eqref{eq:EI_taper} is fit instead, yielding a smooth profile that extrapolates past the poorly conditioned bins.
The direct estimator bins samples by arc length $s$, excluding bins with $s<\SI{0.02}{m}$, where the near-zero-curvature bias is largest.
The remaining bins, pooled across all contact heights, give the taper exponent $p$ via \eqref{eq:p_fit}.
With $p$ fixed at this value, the base flexural-rigidity $EI_0$ is recovered independently via \eqref{eq:EI0_fit}.
Fitting the taper law this way gives, for the twig, $p=0.43$ and $EI_0=2.63\,\text{N\,m}^2$, with a taper-fit $R^2=0.87$.
For the grass, it gives $p=1.60$ and $EI_0=0.112\,\text{N\,m}^2$, with a taper-fit $R^2=0.96$.

\begin{figure}[h!]
    \centering
    \vspace{-0.75em}
    \includegraphics[width=\linewidth]{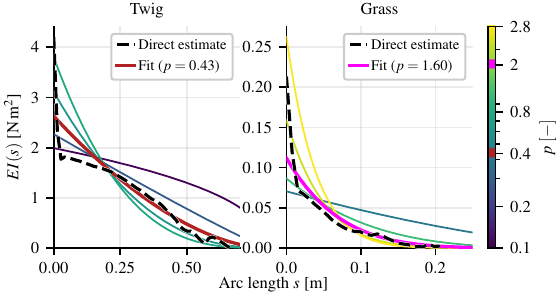}
    \caption{Flexural-rigidity $EI(s)$ against arc length $s$ for the twig (left) and the grass (right). The black dashed curve is the pooled direct estimate. The solid curve is the fitted taper law \eqref{eq:EI_taper}: red for the twig, magenta for the grass. The remaining curves show $EI(s)$ for taper exponents $p$ spanning a range around the fitted value.}
    \label{fig:ei_direct_vs_taper}
    \vspace{-1em}
\end{figure}

\autoref{fig:ei_direct_vs_taper} shows the direct estimate, the fitted taper law, and curves for taper exponents $p$ spanning a range around the fitted value, for both specimens.
The dashed direct estimate spikes sharply near the base for both specimens, consistent with the noise floor argument above.
Apart from the high noise region at the base, the parametrized rigidity matches well the observations.
Further, the twig exhibits higher stiffness, with its rigidity decreasing more gradually from the base than that of the grass, as expected.
Note that, without this parametrization, forward simulations of energy expenditure, stress, or deformation would produce inaccurate results.

A destructive bending test is one way to validate the fitted taper, but its own result depends on load placement \cite{robertson2015measuring}, and is not necessarily more representative of the standing plant than a non-destructive, in-situ estimate \cite{detter2013determining}.
Instead, the fitted taper is used as input to the forward simulation of \eqref{eq:ivp}, and its predictions are evaluated against the measured force and centerline shape, neither of which was included in the fitting procedure.
For each engaged frame, the force error is computed as the difference between the predicted and measured force at the contact point.
The shape error is computed as the RMS distance between the predicted and measured centerlines over up to $N=200$ resampled nodes below the contact point.
The resulting errors are then aggregated over all engaged frames.
At the fitted taper, the twig's median shape RMSE and force error are \SI{14.3}{mm} and \SI{0.47}{N}, relative to its \SI{0.80}{m} length and \SI{5.3}{N} peak force.
The grass's median shape RMSE and force error are \SI{3.9}{mm} and \SI{0.85}{N}, relative to its \SI{0.47}{m} length and \SI{4.1}{N} peak force.

The mechanical profile captures an intrinsic property of the stem, as it is consistent with the direct measurements used for fitting while also predicting force and shape measurements that were not included in the estimation process.
This demonstrates that \eqref{eq:ivp} can be used to predict the force or deflection resulting from a push at a new contact height, without requiring a new fit for each height.
Such predictions can support robotic tasks such as traversability assessment and safe crop interaction.

\subsection{Lumped Rotational Stiffness}\label{sec:ktheta_results}

This subsection evaluates the simpler lumped stiffness $k_\theta$ the same way as \autoref{sec:ei_results}, checking how well it performs and where it breaks down.
As shown in the lumped-stiffness branch of \autoref{fig:global_pipeline}, the torque balance of \eqref{eq:torque_balance} can be fit directly from the force sensor alone, without the shape-extraction pipeline.
\begin{figure}[h!]
    \centering
    \vspace{-0.5em}
    \includegraphics[width=\linewidth]{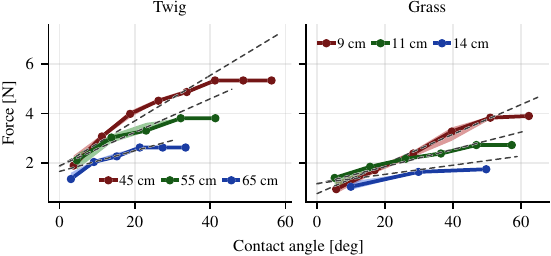}
    \caption{Contact force (median $\pm$ IQR per angle bin) as a function of the deflection angle \(\theta\) (x-axis), with the torque-balance fit (dashed), at three contact heights per specimen. The fitted slope corresponds to the rotational stiffness $k_\theta$ of the lumped-stiffness model.}
    \label{fig:veg_force_fit}
    \vspace{-1em}
\end{figure}

Using only the force measurements from the experiments in \autoref{fig:setup}, \autoref{fig:veg_force_fit} compares the model's predicted force against the measured data, fitting only the linear region of the response.
The stiffness is estimated via \eqref{eq:torque_balance} using only the force sensor, and the deflection angle is computed from \eqref{eq:angle_solution}.
The fitted slope, $k_\theta$, is not the same at each height: it decreases from \SI{0.31}{N.m/rad} to \SI{0.14}{N.m/rad} for the grass, and from \SI{1.60}{N.m/rad} to \SI{0.98}{N.m/rad} for the twig.

After an initial linear region, the measured force reaches a plateau as the vegetation bends sufficiently below the interaction point, allowing the robot to traverse without inducing further deformation.
The torque balance of \eqref{eq:torque_balance} cannot capture this: it assumes one constant stiffness $k_\theta$, so its own prediction keeps rising with $\theta$.

The yield angle determines both the peak force and the elastic work stored in the vegetation up to the onset of yielding, which can be obtained directly from the fitted line.
However, the yield angle depends on the contact point and robot geometry and must therefore be measured separately for each new setup.
Consequently, fitting $k_\theta$ from the force--deflection relationship alone is insufficient to fully characterize the vegetation response.
Unlike the Kirchhoff rod model \eqref{eq:ivp}, the lumped-stiffness model cannot predict the complete mechanical behavior of the plant.
However, when a robot consistently interacts with vegetation at a fixed height and within a limited range of deflections, the lumped-stiffness model remains applicable using only the robot's onboard sensors.
For example, if a robot interacts with crops at a given height without inducing deflections beyond the modeled elastic regime, the unmodeled yielding behavior does not affect the prediction.

\subsection{Generality of the Models}\label{sec:discussion}
This subsection evaluates how well each model's fitted parameters generalize beyond the setup they were fit at, and when that generalization actually matters for a task such as traversability assessment or crop handling.
\begin{figure}[h!]
    \centering
    \vspace{-1.0em}
    \includegraphics[width=\linewidth]{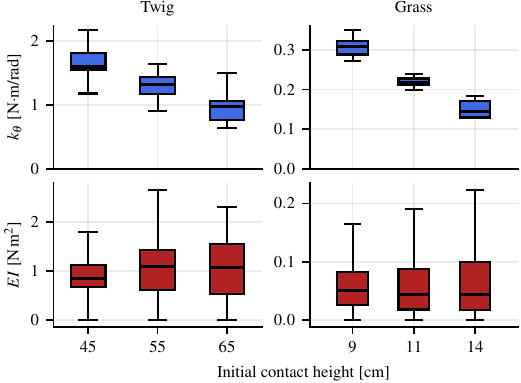}
    \caption{Height-variance of the lumped rotational stiffness $k_\theta$ from the spring-rod model (top) against the distributed flexural rigidity $EI(s)$ from the Kirchhoff-rod model (bottom), both evaluated at the contact height $h$ of each push.}
    \label{fig:height_invariance}
    \vspace{-1em}
\end{figure}

To test this generalization, each specimen was pushed at several different contact heights, and $k_\theta$ and $EI(s)$ were compared across them in \autoref{fig:height_invariance}.
The flexural-rigidity profile $EI(s)$, evaluated at each height from the single pooled profile fit, remain stable across all heights, unlike the lumped stiffness $k_\theta$.
The trend in $k_\theta$ follows directly from \eqref{eq:kdist}.
However, as seen in \autoref{fig:ei_direct_vs_taper}, the rigidity is not constant along the vegetation, and a single parameter cannot capture this complexity.
As such, the results indicate that the flexural-rigidity profile $EI(s)$ provides a representation of the stem's mechanical properties that is independent of the specific interaction used for its estimation.
In contrast, the lumped stiffness $k_\theta$ characterizes the local response of the vegetation at a given contact height and therefore varies with the interaction configuration.

The choice between the two models in \autoref{fig:global_pipeline} depends on whether the interaction geometry remains fixed or varies across the task.
The lumped-stiffness model is sufficient when the robot consistently contacts vegetation at a fixed, known height, as in \textcite{ordonez2020characterization}, who build cost maps from probing vegetation under a fixed contact geometry.
The rod model provides greater flexibility when the contact height varies, as the flexural-rigidity profile $EI(s)$ can be recovered from \eqref{eq:EI_taper} and reused in the forward model of \eqref{eq:ivp} to predict the vegetation response for different contact heights and loads.
This makes the model applicable to tasks involving varying interaction geometries, such as harvesting \cite{deng2024gazebo} or crop-lodging assessment \cite{robertson2016maize}.
Moreover, once the mechanical model has been identified through controlled laboratory experiments, its parameters could be related to observations from exteroceptive sensors.
Building a dataset linking visual observations to the mechanical properties of vegetation is left for future work.

\section{Conclusion}\label{sec:conclusion}
This paper presented a framework for characterizing the vegetation response to interaction.
Two complementary models were evaluated: a spatial flexural-rigidity profile $EI(s)$ estimated from vision and contact-force measurements, and a lumped stiffness $k_\theta$ estimated from force measurements alone.
The experiments showed that $EI(s)$ remains consistent across contact heights, whereas $k_\theta$ depends on the interaction geometry and is applicable only within the small-deformation regime.
The lumped model therefore provides a simple characterization requiring only the robot pose and force measurements, while the spatial model offers greater transferability across contact heights at the cost of additional visual sensing.
Once identified, $EI(s)$ can be reused across different robots, sensing configurations, and contact heights without requiring additional mechanical characterization.
Future work will extend the experimental validation to a broader range of vegetation types and investigate the creation of a dataset linking exteroceptive visual observations and the mechanical properties identified through physical interaction.
The resulting dataset could be used to learn the mechanical response of vegetation from visual observations, enabling its prediction in real-time without force measurements.

\enlargethispage{\baselineskip}



\section*{ACKNOWLEDGMENT}
\footnotesize{This work was funded by the French National Research Agency (ANR) Junior Research Chair program and the International Research Center “Innovation Transportation and Production Systems” of the I-SITE CAP 20-25.
We also thank Nouriddin Asfour and Reine Tindano for their help and support.}
\section*{REFERENCES}

\renewcommand*{\bibfont}{\footnotesize}
\printbibliography[heading=none]
\end{document}